\documentclass{article} 
\usepackage{natbib}
\usepackage{macros}
\usepackage{authblk}
\usepackage{multirow}
\usepackage{natbib}
\usepackage{times}
\usepackage{helvet}
\usepackage{courier}
\pgfplotsset{compat=newest}
\tikzset{roads/.style={line width=0.2cm}}

\newtheorem{theorem}{Theorem}

\newtheorem{proposition}{Proposition}

\title{Policy Gradients for CVaR-Constrained MDPs}
\author{Prashanth L A\thanks{prashanth.la@inria.fr}}
\affil{
INRIA Lille - Nord Europe, Team SequeL, FRANCE.}

\begin{document}
\maketitle


\begin{abstract} 
We study a risk-constrained version of the stochastic shortest path (SSP) problem, where the risk measure considered is Conditional Value-at-Risk (CVaR). We propose two algorithms that obtain a locally risk-optimal policy by employing four tools: stochastic approximation, mini batches, policy gradients and importance sampling. Both the algorithms incorporate a CVaR estimation procedure, along the lines of \cite{bardou2009computing}, which in turn is based on Rockafellar-Uryasev's representation for CVaR and utilize the likelihood ratio principle for estimating the gradient of the sum of one cost function (objective of the SSP) and the gradient of the CVaR of the sum of another cost function (in the constraint of SSP). The algorithms differ in the manner in which they approximate the CVaR estimates/necessary gradients - the first algorithm uses stochastic approximation, while the second employ mini-batches in the spirit of Monte Carlo methods. We establish asymptotic convergence of both the algorithms. 
Further, since estimating CVaR is related to rare-event simulation, we incorporate an importance sampling based variance reduction scheme into our proposed algorithms. 
\end{abstract} 


\section{Introduction}
\label{sec:introduction}

Risk-constrained Markov decision processes (MDPs) have attracted a lot of attention recently in the reinforcement learning (RL) community (cf. \cite{borkar2010risk,tamar2012policy,Prashanth13AC,Tamar13VA}). However, unlike previous works 
that focused mostly on variance of the return as a measure of risk, we consider Conditional Value-at-Risk (CVaR) as a risk measure.
CVaR has the form of a conditional expectation, where the conditioning is based on a constraint on Value-at-Risk (VaR).

The aim in this paper is to find a {\em risk-optimal} policy in the context of a stochastic shortest path (SSP) problem. A risk-optimal policy is one that minimizes the sum of one cost function (see $G^{\theta}(s^0)$ in \eqref{eq:cvar-ssp}), while ensuring that the conditional expectation of the sum of another cost function (see $C^{\theta}(s^0)$ in \eqref{eq:cvar-ssp}) given some confidence level, stays bounded, i.e., the solution to the following  risk-constrained problem: For a given $\alpha \in (0,1)$ and $K_\alpha >0$, 
\begin{align}
\label{eq:cvar-ssp}
\min_{\theta\in \Theta} \underbrace{\E\left[ \sum\limits_{m=0}^{\tau-1} g(s_m,a_m) \left| s_0=s^0 \right. \right]}_{G^{\theta}(s^0)}   \text{~subject to~}  \cvar\underbrace{\left[ \sum\limits_{m=0}^{\tau-1} c(s_m,a_m)\left|s_0=s^0 \right.\right]}_{C^{\theta}(s^0)}\leq K_\alpha.
\end{align}
In the above, $s^0$ is the starting state and the actions $a_0,\ldots,a_{\tau-1}$ are chosen according to a randomized policy $\pi_\theta$ governed by $\theta$. Further, $g(s,a)$ and $c(s,a)$ are cost functions that take a state $s$ and an action $a$ as inputs and $\tau$ is the first passage time to the recurrent state of the underlying SSP (see Section \ref{sec:problem} for a detailed formulation). In \cite{borkar2010risk}, a similar problem is considered in a finite horizon MDP, though under a strong separability assumption for the cost function $c(s,a)$.    

Solving the risk-constrained problem \eqref{eq:cvar-ssp} is challenging due to two reasons:\\
\begin{inparaenum}[\bfseries(i)]
\item Finding a globally risk-optimal policy is intractable even for a simpler case when the risk is defined as the variance of the return of an MDP (see \cite{mannor2011mean}). The risk-constrained MDP that we consider is more complicated in comparison, since CVaR is a conditional expectation, with the conditioning governed by an event that bounds a probability.\\ 
\item For the sake of optimization of the CVaR-constrained MDP that we consider in this paper, it is required to estimate both VaR/CVaR of the total cost ($C^{\theta}(s^0)$ in \eqref{eq:cvar-ssp}) as well as its gradient. The problem is further complicated by the fact  VaR/CVaR concerns the tail of the distribution of the total cost and hence, a variance reduction technique is required to speed up the estimation procedure. \\
\end{inparaenum}
We avoid the first problem by proposing a policy gradient scheme that is proven to converge to a locally optimal policy, while the second problem is alleviated using two principled approaches: stochastic approximation/mini-batch procedure for estimating VaR/CVaR derived out of a well-known convex optimization problem by \\\cite{rockafellar2000optimization} and likelihood ratio estimates from the classic policy-gradient algorithm by \\\cite{bartlett2011infinite}. 

The contributions of this paper are summarized as follows:\\
\begin{inparaenum}[\bfseries(I)]
\item 
First, 
using the representation of CVaR (and also VaR) as the solution of a certain convex optimization problem by \cite{rockafellar2000optimization}, we develop a stochastic approximation procedure, along the lines of \cite{bardou2009computing}, for estimating the CVaR  of a policy for an SSP. 
In addition, we also propose a scheme based on the mini-batch principle to estimate CVaR. Mini-batches are in the spirit of Monte Carlo methods and have been proposed by \cite{atchade2014stochastic} under a different optimization context for stochastic proximal gradient algorithms. \\
\item Second, we develop two novel policy gradient algorithms for finding a (locally) risk-optimal policy of the CVaR-constrained SSP. The first algorithm is a four time-scale stochastic approximation scheme while the second operates along two time-scales in conjunction with mini-batches.  Both algorithms use a procedure to estimate CVaR and then use the policy-gradient principle with likelihood ratios to estimate the gradient of the total cost $G^{\theta}(s^0)$ as well as CVaR of another cost sum $C^{\theta}(s^0)$. Using the CVaR estimates as well as the necessary gradients (estimated along the fastest two time-scales), the first algorithm updates the policy parameter in the negative descent direction on the intermediate timescale and performs dual ascent for the Lagrange multiplier on the slowest timescale. On the other hand, the second algorithm operates on two timescales as it employs mini-batches to estimate the CVaR as well as the necessary gradients. \\
\item Third, we adapt our proposed algorithms to incorporate importance sampling (IS) -  a well-known variance-reduction scheme. This is motivated by the fact that when the confidence level $\alpha$ is close to $1$, estimating VaR as well as CVaR takes longer. This is because the interesting samples used to estimate CVaR come from the tail of the distribution of the random variable  concerned (in our case, the total cost $C^{\theta}(s^0)$) and thus, get rarer as $\alpha$ gets close to $1$. Importance sampling (IS) is a standard tool to alleviate this problem and we employ the IS scheme proposed by \cite{lemaire2010unconstrained}.
However, applying the latter scheme in a SSP setting is non-trivial as it requires the knowledge of transition dynamics. We propose a heuristic IS variant where we use the randomized policies to derive sampling ratios for the IS procedure. \\
\end{inparaenum}
To sum up, the core contribution of this paper is twofold. First, using a careful synthesis of well-known techniques from stochastic approximation, likelihood ratios and importance sampling, we propose a policy gradient algorithm that is provably convergent to a locally risk-optimal policy. Second, we propose another algorithm based on the idea of mini-batches for estimating CVaR from the simulated samples. The latter approach is novel even for policy gradients in the context of risk-neutral MDPs.

The rest of the paper is organized as follows:  In Section \ref{sec:problem} we formalize the CVaR-constrained SSP and in Section \ref{sec:structure} describe the structure of our proposed algorithms. In Section \ref{sec:pgcvarsa} we present the first algorithm based on stochastic approximation and in Section \ref{sec:pgcvarmb} we present the mini-batch variant. In Section \ref{sec:convergence}, we sketch the convergence of our algorithms and later in Section \ref{sec:is} describe the importance sampling variants of our algorithms. In Section \ref{sec:related-work}, we review relevant previous works. Finally, in Section \ref{sec:conclusions} we provide the concluding remarks.

\section{Problem formulation}
\label{sec:problem}
In this section, we first introduce VaR/CVaR risk measures, then formalize the stochastic shortest path problem and subsequently define the CVaR-constrained SSP.
\subsection{Background on VaR and CVaR}
For any random variable $X$, we define the VaR at level $\alpha\in\left(0,1\right)$ as
$$
\text{VaR}_{\alpha}(X):=\inf\left\{\xi \mbox{ }|\mbox{ }\mathbb{P}\left(X\leq \xi\right)\geq\alpha\right\}.
$$
If the distribution of $X$ is continuous, then VaR is the lowest solution to $\mathbb{P}\left(X\leq \xi\right)=\alpha.$
VaR as a risk measure has several drawbacks, which precludes using standard stochastic optimization methods. This motivated the definition of coherent risk measures by \cite{artzner1999coherent}. A risk measure is coherent if it is convex, monotone, positive homogeneous and translation equi-variant. CVaR is one popular risk measure  defined by
$$
\text{CVaR}_{\alpha}(X):=\mathbb{E}\left[X | X \geq \text{VaR}_{\alpha}(X)\right].
$$
Unlike VaR, the above is a coherent risk measure.

\subsection{Stochastic Shortest Path (SSP)}
We consider a SSP with a finite state space $\S=\{0,1,\ldots,r\}$, where $0$ is a special cost-free terminal state. The set of feasible actions in state $s \in \S$ is denoted by $\A(s)$. A transition from state $s$ to $s'$ under action $a \in \A(s)$ occurs with probability $p_{ss'}(a)$ and incurs the following costs: $g(s,a)$ and $c(s,a)$, respectively. The terminal state $0$ is cost-free and absorbing. 

A policy specifies how actions are chosen in each state. A {\em stationary} randomized policy $\pi(\cdot|s)$ maps any state $s$ to a probability vector on $\A(s)$. As is standard in policy gradient algorithms, we parameterize the policy and assume that the policy is continuously differentiable in the parameter $\theta$. Since a policy $\pi$ is identifiable by its parameter $\theta$, we use them interchangeably in this paper. 

As defined by \cite{bertsekas1995dynamic}, a proper policy is one which ensures that there is a positive probability that the terminal state $0$ will be reached, starting from any initial state, after utmost $r$ transitions. This in turn implies the states $1,\ldots,r$ are transient. We assume that class of parameterized policies considered, i.e., $\{\pi_\theta \mid \theta \in \Theta\}$, is proper. 

\subsection{CVaR-constrained SSP}
As outlined earlier, the risk-constrained objective is: 
\begin{align*}
\hspace{-0.4em}\min_{\theta\in \Theta} \underbrace{\E\left[ \sum\limits_{m=0}^{\tau-1} g(s_m,a_m) \left| s_0=s^0 \right. \right]}_{G^{\theta}(s^0)}   \text{~subject to~}  \cvar\underbrace{\left[ \sum\limits_{m=0}^{\tau-1} c(s_m,a_m)\left|s_0=s^0 \right.\right]}_{C^{\theta}(s^0)}\leq K_\alpha,
\end{align*}
where $\tau$ denotes the first visiting time to terminal state $0$, i.e., $\tau = \min \{m \mid s_m = 0\}$. The actions $a_0,\ldots,a_{\tau-1}$ are chosen according to the randomized policy $\pi_\theta$. Further, $\alpha$ and $K_\alpha$ are constants that specify the confidence level and constraint bound for CVaR, respectively.

Using the standard trick of Lagrangian relaxation for constrained optimization problems, we convert \eqref{eq:cvar-ssp} to the following unconstrained problem:  
\begin{equation}
\label{eq:unconstrained-cvar-ssp}
\max_\lambda\min_\theta\left[\L^{\theta,\lambda}(s^0) := G^{\theta}(s^0)+\lambda\big(\cvar(C^{\theta}(s^0))-K_\alpha\big)\right].
\end{equation}

\section{Algorithm Structure}
\label{sec:structure}
In order to solve \eqref{eq:unconstrained-cvar-ssp}, a standard constrained optimization procedure operates as follows:
  \begin{description}
   \item[Simulation.] This is the inner-most loop where the SSP is simulated for several episodes and the resulting costs are aggregated.  
   \item[Policy Update.]  This is the intermediate loop where the gradient of the Lagrangian along $\theta$ is estimated using simulated values above. The gradient estimates are then used to update policy parameter $\theta$ along a descent direction. Note that this loop is for a given value of $\lambda$; and
   \item[Lagrange Multiplier Update.] This is the outer-most loop where the Lagrange multiplier $\lambda$ is updated along an ascent direction, using the converged values of the inner two loops.
  \end{description}

\begin{figure}
\centering
\tikzstyle{block} = [draw, fill=white, rectangle,
   minimum height=5em, minimum width=6em]
\tikzstyle{sum} = [draw, fill=white, circle, node distance=1cm]
\tikzstyle{input} = [coordinate]
\tikzstyle{output} = [coordinate]
\tikzstyle{pinstyle} = [pin edge={to-,thin,black}]
\scalebox{0.8}{\begin{tikzpicture}[auto, node distance=2cm,>=latex']
\node (theta) at (-10,0) {$\bm{\theta_n}$};
\node [block, fill=blue!20,right=2cm of theta,label=below:{\color{bleu2}\bf Simulation},align=center] (sample) {\textbf{Using policy $\pi_{\theta_n}$, }\\[1ex] \textbf{simulate an SSP episode}}; 
\node [block, fill=green!20,above right=3cm of sample,label=below:{\color{darkgreen!90}\bf Policy Gradient},align=center] (obj) {\textbf{Estimate} $\bm{\nabla_\theta G^{\theta}(s^0)}$};
\node [block, fill=green!20,right=2cm of sample,label=below:{\color{darkgreen!90}\bf CVaR Estimation},align=center] (estcvar) {\textbf{Estimate} $\bm{\cvar(C^{\theta}(s^0))}$};
\node [block, fill=green!20,below right=3cm of sample,label=below:{\color{darkgreen!90}\bf CVaR Gradient},align=center] (cvar) {\textbf{Estimate}\\ $\bm{\nabla_\theta \cvar(C^{\theta}(s^0))}$};
\node [block, fill=red!20,right=8cm of sample, minimum height=8em,label=below:{\color{violet!90}\bf Policy Update},align=center] (update) {\textbf{Update} $\bm{\theta_n}$ \\[1ex]\textbf{using \eqref{eq:pgcvarsa} or \eqref{eq:pgcvarmb}}};
\node [right=2cm of update] (end) {$\bm{\mathbf{\theta_{n+1}}}$};
\draw [thick,-triangle 45] (theta) --  (sample);
\draw [thick,-triangle 45] (sample) -- (obj);
\draw [thick,-triangle 45] (sample) -- (estcvar);
\draw [thick,-triangle 45] (sample) -- (cvar);
\draw [thick,-triangle 45] (obj) -- (update);
\draw [thick,-triangle 45] (estcvar) -- (update);
\draw [thick,-triangle 45] (cvar) -- (update);
\draw [thick,-triangle 45] (estcvar) -- (cvar);
\draw [thick,-triangle 45] (update) -- (end);
\end{tikzpicture}}
\caption{Overall flow of our algorithms.}
\label{fig:algorithm-flow}
\end{figure}
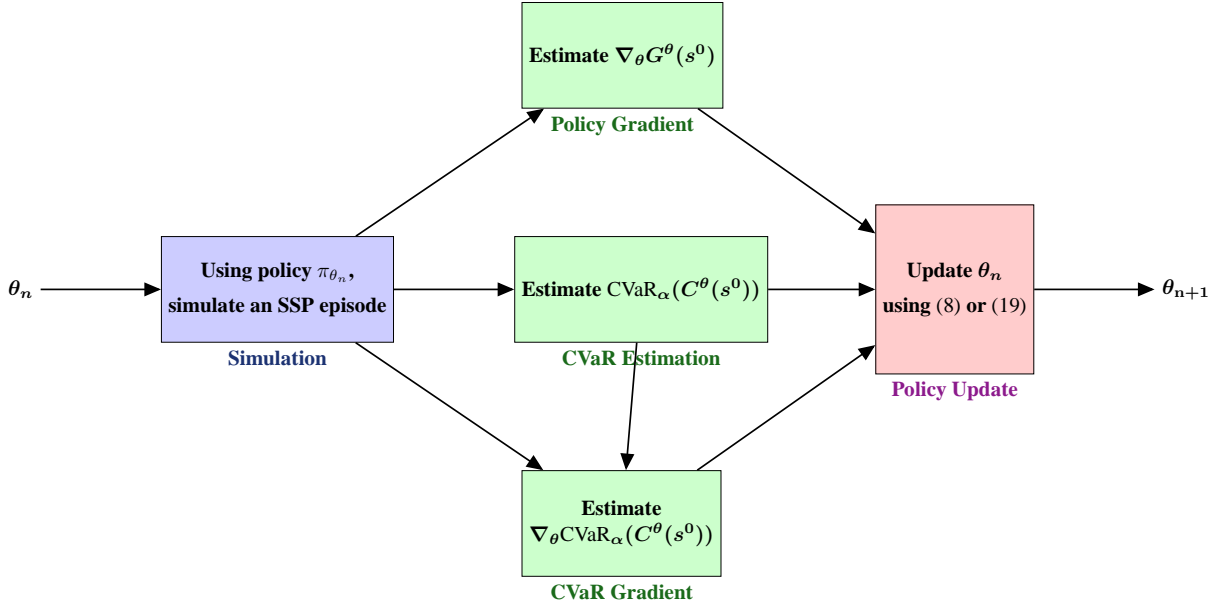

Using two-timescale stochastic approximation (see Chapter $6$ of \cite{borkar2008stochastic}), the policy and Lagrange multiplier update can run in parallel as follows:
\begin{align}
\label{eq:theta-update-general}
\theta_{n+1} &= \theta_n - \gamma_n \nabla_\theta \L^{\theta,\lambda}(s^0) \text{~~~ and ~~~}
\lambda_{n+1} = \Gamma_\lambda\big[\lambda_{n} + \beta_n \nabla_\lambda \L^{\theta,\lambda}(s^0)\big],
\end{align}
where $\Gamma_\lambda$ is a projection operator that keep the iterate $\lambda_n$ bounded, while $\gamma_n,\beta_n, n\ge 0$ are step-sizes that satisfy the following assumption:
  \[\begin{array}{l}
\sum\limits_{n = 1}^\infty \beta_n = \infty,  \sum\limits_{n = 1}^\infty \gamma_n = \infty,
\sum\limits_{n = 1}^\infty \left ( \beta_n^2 + \gamma_n^2 \right ) < \infty \text{ and } \dfrac{\beta_n}{\gamma_n} \rightarrow 0.
  \end{array}
\]
The last condition above ensures that $\theta$-recursion proceeds on a faster timescale in comparison to $\lambda$-recursion.\\[1ex]
\noindent\textbf{Simulation optimization.} No closed form expression for the gradient of the Lagrangian $L^{\theta,\lambda}(s^0)$ is available and moreover, $G^\theta(s^0)$ and $C^\theta(s^0)$ are observable only via simulation. Observe that $\nabla_\theta \L^{\theta,\lambda}(s^0)= \nabla_\theta G^{\theta}(s^0) + \lambda \nabla_\theta \cvar(C^{\theta}(s^0))$ and $\nabla_\lambda \L^{\theta,\lambda}(s^0)= \cvar(C^{\theta}(s^0))-K_\alpha$. 
Hence, in order to update according to \eqref{eq:theta-update-general}, we need to estimate, for any policy parameter $\theta$, the following quantities via simulation:\\ 
\begin{inparaenum}[\bfseries(i)]
\item $\cvar(C^{\theta}(s^0))$;~~~~~~~
\item $\nabla_\theta G^{\theta}(s^0)$;~~~~~~~ and ~~~~~~ 
\item $\nabla_\theta\cvar(C^{\theta}(s^0))$.\\                                                                                                                                                                 \end{inparaenum}
In the following sections, we describe two algorithms that differ in the way they estimate each of the above quantities and subsequently establish that the estimates (and hence the overall algorithms) converge.

\section{Algorithm 1: PG-CVaR-SA}
\label{sec:pgcvarsa}
Algorithm \ref{alg:pgcvarsa} describes the complete algorithm along with the update rules for the various parameters.
The algorithm involves the following crucial components -  simulation of the SSP, VaR/CVaR estimation and policy gradients for the objective as well as the CVaR constraint. Each of these components is described in detail in the following. 

\begin{algorithm}[h]
\caption{PG-CVaR-SA}
\label{alg:pgcvarsa}
\begin{algorithmic}
\State {\bfseries Input:} parameterized policy $\pi_\theta(\cdot|\cdot)$, step-sizes $\{\zeta_{n,1},\zeta_{n,2}, \gamma_n, \beta_n\}_{n \ge 1}$
\State {\bfseries Initialization:} Starting state $s^0$, initial policy $\theta_0$, , number of iterations $M >> 1$. 
\For {$n=1,2,\ldots,M$}
\State \textbf{Simulation}
\begin{flalign*}
\hspace{2em}&\text{Simulate the SSP for an episode using actions }a_{n,0},\ldots,a_{n,\tau_n-1}\text{ generated using }\pi_{\theta_{n-1}}&\\[0.5ex]
\hspace{2em}&\text{Obtain cost estimates:~~}G_n:=\sum\limits_{j=0}^{\tau_n-1} g(s_{n,j},a_{n,j})\text{ and }C_n:=\sum\limits_{j=0}^{\tau_n-1} c(s_{n,j},a_{n,j})& \\[0.5ex]
\hspace{2em}&\text{Obtain likelihood derivative:~~} z_n:=\sum\limits_{j=0}^{\tau_n-1} \nabla \log \pi_\theta(s_{n,j},a_{n,j})&
\end{flalign*}
\State \textbf{VaR/CVaR estimation:}
\begin{flalign}
&\hspace{2em}\text{VaR:~~~~~ }      \xi_{n}= \xi_{n-1}-\zeta_{n,1}\left(1-\frac{1}{1-\alpha}\mbox{\bf 1}_{\left\{C_n\geq\xi_{n-1}\right\}}\right),&\\
&\hspace{2em}\text{CVaR:~~~ }     \psi_{n}=\psi_{n-1}-\zeta_{n,2}\left(\psi_{n-1} - v(\xi_{n-1},C_{n})\right).&
\end{flalign}
\State \textbf{Policy Gradient:}
\begin{flalign}
&\hspace{2em}\text{Total Cost:~ }      \bar G_{n}= \bar G_{n-1}-\gamma_{n}(G_n - \bar G_{n}),
\hspace{0.7em}\text{Gradient:~ }     \partial G_n =  \bar G_{n} z_n.\label{eq:pg-grad-sa}&
\end{flalign}
\State \textbf{CVaR Gradient:}
\begin{flalign}
&\hspace{2em}\text{Total Cost:~ }      \tilde C_{n}= \tilde C_{n-1}-\gamma_{n}(C_n - \tilde C_{n}),
\hspace{0.7em}\text{Gradient:~ }     \partial C_n =  (\tilde C_{n} - \xi_n) z_n \mbox{\bf 1}_{\left\{C_n\geq\xi_{n}\right\}}.\label{eq:cvar-grad-sa}&
\end{flalign}
\State \textbf{Policy and Lagrange Multiplier Update:}
\begin{flalign}
&\hspace{2em}\theta_n = \theta_{n-1}-\gamma_{n}(\partial G_n + \lambda_{n-1} (\partial C_n)),
\hspace{1.7em}\lambda_{n}=\Gamma_\lambda\Big(\lambda_{n-1}+\beta_n(\psi_n-K_\alpha)\Big). \label{eq:pgcvarsa}
\end{flalign}
\EndFor
\State Output $(\theta_M,\lambda_M)$.
\end{algorithmic}
\end{algorithm}

\subsection{SSP Simulation}
In each iteration of PG-CVaR-SA, an episode of the underlying SSP is simulated. Each episode ends with a visit to the recurrent state $0$ of the SSP. Let $\tau_n$ denote the time of this visit in episode $n$. The actions $a_{n,j}, j=0,\ldots,\tau_n-1$ in episode $n$ are chosen according to the policy $\pi_{\theta_{n-1}}$.  Let $G_n:=\sum\limits_{j=0}^{\tau_n-1} g(s_{n,j},a_{n,j})$ and $C_n:=\sum\limits_{j=0}^{\tau_n-1} c(s_{n,j},a_{n,j})$ denote the accumulated cost values. Further, let $z_n:=\sum\limits_{j=0}^{\tau_n-1} \nabla \log \pi_\theta(s_{n,j},a_{n,j})$ denote the likelihood derivative (see Section \ref{sec:pg} below). 
The tuple $(G_n,C_n,z_n)$ obtained at the end of the $n$th episode is used to estimate CVaR as well as policy gradients.

\subsection{Estimating VaR and CVaR}
A well-known result by \cite{rockafellar2000optimization} is that both VaR and CVaR can be obtained from the solution of a certain convex optimization problem and we recall this result next.  
\begin{theorem}
\label{thm:var-cvar-convex}
For any random variable $X$, let 
\begin{align}
\label{eq:vV}
v(\xi,X):=\xi + \frac{1}{1-\alpha}(X-\xi)_{+}  \text{ and } V(\xi)=\E\left[v(\xi,X)\right]
\end{align}
Then, $\textnormal{VaR}_{\alpha}(X)$ is any point of the set $\arg\min V= {\left\{\xi \in \mathbb{R}\ | \ V'(\xi)=0 \right\}}$, 
where $V'$ is the derivative of $V$ w.r.t. $\xi$.
Further, $\textnormal{CVaR}_{\alpha}(X)=V(\xi^{*}_{\alpha})$, where $\xi^{*}_{\alpha}$ is a $\textnormal{VaR}_{\alpha}(X)$. 
\end{theorem}

From the above, it is clear that in order to estimate VaR/CVaR, one needs to find a $\xi$ that satisfies $V'(\xi)=0$. Stochastic approximation (SA) is a natural tool to use in this situation. We briefly introduce SA next and later develop a scheme for estimating CVaR along the lines of \cite{bardou2009computing} on the faster timescale of PG-CVaR-SA.\\[1ex]
\noindent\textbf{Stochastic approximation.}
The aim is to solve the equation $F(\theta)  = 0$ when analytical form of $F$ is
not known. However, noisy measurements $F(\theta_n) + \xi_n$ can be
obtained, where $\theta_n, n \ge 0$ are the input parameters and $\xi_n, n \ge 0$ are zero-mean random variables, that are not necessarily i.i.d. 

The seminal Robbins Monro algorithm solved this problem by employing the following update rule:
\begin{align}
\label{eq:sa}
\theta_{n+1} = \theta_n + \gamma_n (F(\theta_n)+\xi_n).
\end{align}
In the above, $\gamma_n$ are step-sizes that satisfy $\sum\limits_{n = 1}^\infty \gamma_n = \infty$ and $\sum\limits_{n = 1}^\infty \gamma_n^2  < \infty$. Under a stability assumption for the iterates and bounded noise, it can be shown that  $\theta_n$ governed by \eqref{eq:sa} converges to the solution of   $F(\theta)  = 0$ (cf. Proposition \ref{prop:lyapunov} in Section \ref{sec:convergence}).\\[1ex]
\subsubsection{CVaR estimation using SA.}
Using the stochastic approximation principle and the result in Theorem \ref{thm:var-cvar-convex}, we have the following scheme to estimate the VaR/CVaR simultaneously from the simulated samples $C_n$:
\begin{align}
\text{VaR: } &     \xi_{n}= \xi_{n-1}-\zeta_{n,1}(\underbrace{1-\frac{1}{1-\alpha}\mbox{\bf 1}_{\left\{C_n\geq\xi\right\}}}_{\frac{\partial v}{\partial \xi}(\xi,C_n)}),
\label{eq:RMVaR}\\
\text{CVaR: } &    \psi_{n}=\psi_{n-1}-\zeta_{n,2}\left(\psi_{n-1} - v(\xi_{n-1},C_{n})\right).
\label{eq:RMCVaR}
\end{align}
In the above, \eqref{eq:RMVaR} can be seen as a gradient descent rule, while \eqref{eq:RMCVaR} can be seen as a plain averaging update.
The scheme above is similar to the one proposed by \cite{bardou2009computing}, except that the random variable $C^\theta(s^0)$ (whose CVaR we try to estimate) is a sum of costs obtained at the end of each episode, unlike the single-shot r.v. considered by \cite{bardou2009computing}.
The step-sizes $\zeta_{n,1}, \zeta_{n,2}$ satisfy
  \[\begin{array}{l}
\sum\limits_{n = 1}^\infty \zeta_{n,1} = \infty,  \sum\limits_{n = 1}^\infty \zeta_{n,2} = \infty,
\sum\limits_{n = 1}^\infty \left ( \zeta_{n,1}^2 + \zeta_{n,2}^2 \right ) < \infty, \dfrac{\zeta_{n,2}}{\zeta_{n,1}} \rightarrow 0 \text{ and } \dfrac{\gamma_n}{\zeta_{n,2}} \rightarrow 0.
  \end{array}
\]
The last two conditions above ensure that VaR estimation recursion \eqref{eq:RMVaR} proceeds on a faster timescale in comparison to CVaR estimation recursion \eqref{eq:RMCVaR} and further, the CVaR recursion itself proceeds on a faster timescale as compared to the policy parameter $\theta$-recursion. 

Using the ordinary differential equation (ODE) approach, we establish later that the tuple $(\xi_n,\psi_n)$ converges to $\var(C^\theta(s^0)),\cvar(C^\theta(s^0))$, for any fixed policy parameter $\theta$ (see Section \ref{sec:convergence}).

\subsection{Policy Gradient}
\label{sec:pg}
We briefly introduce the technique of likelihood ratios for gradient estimation \cite{glynn1987likelilood} and later provide the necessary estimate for the gradient of total cost $G^\theta(s^0)$.\\[1ex] 
\subsubsection{Gradient estimation using likelihood ratios.}
Consider a Markov chain $\{X_n\}$ with a single recurrent state $0$ and transient states $1,\ldots,r$. Let $P(\theta) := [[p_{X_i X_{j}}(\theta)]]_{i,j=0}^{r}$ denote the transition probability matrix of this chain. Here $p_{X_i X_{j}}(\theta)$ denotes the probability of going from state $X_i$ to $X_{j}$ and is parameterized by $\theta$. Let $\tau$ denote the first passage time to the recurrent state $0$.

Let $X:=(X_0, \ldots, X_{\tau-1})^T$ denote the sequence of states encountered between visits to the recurrent state $0$.
The aim is to optimize a performance measure $F(\theta)=\E[f(\theta,X)]$ for this chain using simulated values of $X$.  
The likelihood estimate is obtained by first simulating the Markov chain according to $P(\theta)$ to obtain the samples $X_0, \ldots, X_{\tau-1}$ and then estimate the gradient as follows:
$$\nabla_\theta F(\theta) = \E \left[ f(X) \sum\limits_{m=0}^{\tau-1} \dfrac{\nabla_\theta p_{X_m X_{m+1}}(\theta)}{p_{X_m X_{m+1}}(\theta)} \right].$$

\subsubsection{Policy Gradient for the objective.}
For estimating the gradient of the objective $G^\theta(s^0)$, we employ the following well-known estimate (cf. \cite{bartlett2011infinite}):
\begin{align}
 \nabla_\theta G^\theta(s^0) = \E\left[ \left(\sum\limits_{n=0}^{\tau-1} g(s_n,a_n)\right) \nabla \log P(s_0,\ldots,s_{\tau-1}) \mid s_0=s^0 \right],
\end{align}
where $\nabla \log P(s_0,\ldots,s_\tau)$ is the likelihood derivative for a policy parameterized by $\theta$, defined as
\begin{align}
\label{eq:pgrad}
 \nabla \log P(s_0,\ldots,s_{\tau-1}) = \sum\limits_{m=0}^{\tau-1} \nabla \log \pi_\theta(a_m\left|s_m\right.).
\end{align}
The above relation holds owing to the fact that we parameterize the policies and hence, the gradient of the transition probabilities can be estimated from the policy alone. This is the well-known policy gradient technique \cite{bartlett2011infinite} that makes it amenable for estimating gradient of a performance measure in MDPs, since the transition probabilities are not required and one can work with policies and simulated transitions from the MDP.

\subsection{Policy Gradient for the CVaR constraint.}
For estimating the gradient of the CVaR of $C^\theta(s^0)$ for a given policy parameter $\theta$, we employ the following likelihood estimate proposed by \cite{tamar2014policy}:
\begin{align}
 &\nabla_\theta \cvar(C^\theta(s^0))  \label{eq:cvargrad} \\
 &= \E\left[ \left( C^\theta(s^0)- \var(C^\theta(s^0)) \right) \nabla \log P(s_0,\ldots,s_{\tau-1}) \mid C^\theta(s^0) \geq \text{VaR}_{\alpha}(C^\theta(s^0)) \right],\nonumber
\end{align}
where $\nabla \log P(s_0,\ldots,s_\tau)$ is as defined before in \eqref{eq:pgrad}. 

Since we do not know $\var(C^\theta(s^0))$, in Algorithm \ref{alg:pgcvarsa} we have an online scheme that uses $\xi_n$ (see \eqref{eq:RMVaR}) to approximate $\var(C^\theta(s^0))$, which is then used to derive an approximation to the gradient $\nabla_\theta \cvar(C^\theta(s^0))$ (see \eqref{eq:cvar-grad-sa}).

\section{Algorithm 2: PG-CVaR-mB}
\label{sec:pgcvarmb}
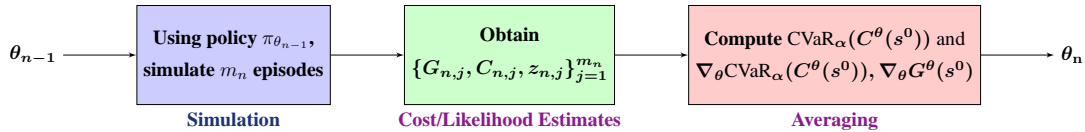
\begin{figure}
\centering
\tikzstyle{block} = [draw, fill=white, rectangle,
   minimum height=5em, minimum width=6em]
\tikzstyle{sum} = [draw, fill=white, circle, node distance=1cm]
\tikzstyle{input} = [coordinate]
\tikzstyle{output} = [coordinate]
\tikzstyle{pinstyle} = [pin edge={to-,thin,black}]
\scalebox{0.75}{\begin{tikzpicture}[auto, node distance=2cm,>=latex']
\node (theta) {$\boldsymbol{\theta_{n-1}}$};
\node [block, fill=blue!20,right=1.3cm of theta,label=below:{\color{bleu2}\bf Simulation},align=center] (sample) {\textbf{Using policy $\pi_{\theta_{n-1}}$, }\\[1ex] \textbf{simulate $m_n$ episodes}}; 
\node [block, fill=green!20,right=1.3cm of sample,label=below:{\color{violet!90}\bf Cost/Likelihood Estimates},align=center] (simval) {\textbf{Obtain} \\[1ex]\textbf{$\boldsymbol{\{G_{n,j}, C_{n,j},z_{n,j}\}_{j=1}^{m_n}}$}};
\node [block, fill=red!20,right=1.3cm of simval,label=below:{\color{violet!90}\bf Averaging},align=center] (update) {\textbf{Compute} $\boldsymbol{\cvar(C^\theta(s^0))}$ and \\[1ex]$\boldsymbol{\nabla_\theta \cvar(C^\theta(s^0)), \nabla_\theta G^\theta(s^0)}$};
\node [right=1.3cm of update] (end) {$\boldsymbol{\mathbf{\theta_{n}}}$};
\draw [->] (theta) --  (sample);
\draw [->] (sample) -- (simval);
\draw [->] (simval) -- (update);
\draw [->] (update) -- (end);
\end{tikzpicture}}
\caption{Illustration of mini-batch principle in PG-CVaR-mB algorithm.}
\label{fig:algorithm-flow-mb}
\end{figure}
\noindent\textbf{mini-Batches.} As illustrated in Figure \ref{fig:algorithm-flow-mb}, we simulate the SSP for several episodes in each iteration of PG-CVaR-mB. 
At the end of the simulation, we obtain the total costs and likelihood derivative estimates $\{G_{n,j}, C_{n,j},z_{n,j}\}_{j=1}^{m_n}$ and using these,  the following quantities are approximated: $\cvar(C^\theta(s^0))$,$\nabla_\theta \cvar(C^\theta(s^0))$ and $\nabla_\theta G^\theta(s^0)$. The latter approximations can be seen as empirical means of functions of $G_{n,j}, C_{n,j},z_{n,j}$.
The complete algorithm along with the update rules for various parameters is presented in Algorithm \ref{alg:pgcvarmB}.
\begin{algorithm}
\caption{PG-CVaR-mB}
\label{alg:pgcvarmB}
\begin{algorithmic}
\State {\bfseries Input:} parameterized policy $\pi_\theta(\cdot|\cdot)$, step-sizes $\{\gamma_n,\beta_n\}$, non-negative weights $\{a_n\}$, mini-batch sizes $\{m_n\}$.
\State {\bfseries Initialization:} Starting state $s^0$, initial policy $\theta_0$, number of iterations $M>>1$. 
\For {$n=1,2,\ldots,M$}
\State \textbf{Simulation}
\begin{flalign*}
&\hspace{2em}\text{Simulate the SSP for $m_n$  episodes using actions generated from the policy }\pi_{\theta_{n-1}} &\\[0.5ex]
&\hspace{2em}\text{Obtain cost estimates:~~} \{G_{n,j}, C_{n,j}\}_{j=1}^{m_n}&\\[0.3ex]
&\hspace{2em}\text{Obtain likelihood derivatives:~~} \{z_{n,j}\}_{j=1}^{m_n}& 
\end{flalign*}
\State \textbf{VaR/CVaR estimation:}
\begin{flalign}
&\hspace{2em}\text{VaR:~ }      \xi_{n}= \dfrac{1}{m_n} \sum\limits_{j=1}^{m_n} \bigg(1-\frac{\mbox{\bf 1}_{\left\{C_{n,j}\geq\xi_{n-1}\right\}}}{1-\alpha}\bigg),
\hspace{0.8em}\text{CVaR:~ }     \psi_{n}=\dfrac{1}{m_n} \sum\limits_{j=1}^{m_n} v(\xi_{n-1},C_{n,j})\label{eq:cvarmb}&
\end{flalign}
\State \textbf{Policy Gradient:}
\begin{flalign}
&\hspace{2em}\text{Total Cost: }      \bar G_{n}= \dfrac{1}{m_n} \sum\limits_{j=1}^{m_n}G_{n,j},
\hspace{2em}\text{Gradient: }     \partial G_n =  \bar G_n z_n.\label{eq:pg-grad-mb}&
\end{flalign}
\State \textbf{CVaR Gradient:}
\begin{flalign}
&\hspace{2em}\text{Total Cost: }      \bar C_{n}= \dfrac{1}{m_n} \sum\limits_{j=1}^{m_n}C_{n,j},
\hspace{2em}\text{Gradient: }     \partial C_n =  (\tilde C_{n} - \xi_n) z_n \mbox{\bf 1}_{\left\{\bar C_n\geq\xi_{n}\right\}}.\label{eq:cvar-grad-mb}
\end{flalign}
\State \textbf{Policy and Lagrange Multiplier Update:}
\begin{flalign}
&\hspace{2em}\theta_n = \theta_{n-1}-\gamma_{n}(\partial G_n + \lambda_{n-1} (\partial C_n)), 
\hspace{1.6em}\lambda_{n}=\Gamma_\lambda\Big(\lambda_{n-1}+\beta_n(\psi_n-K_\alpha)\Big) \label{eq:pgcvarmb}&
\end{flalign}
\EndFor
\State Output $\left(\bar\theta_M := \dfrac{\sum_{k=1}^M a_k \theta_k}{\sum_{k=1}^M a_k},\lambda_M\right)$.
\end{algorithmic}
\end{algorithm}

\noindent\textbf{mini-Batch size.} A simple setting for the batch-size $m_n$ is $C n^\delta$ for some $\delta > 0$, i.e., $m_n$ increases as a function of $n$. We cannot have constant batches, i.e., $\delta = 0$, since the bias of the CVaR estimates and the gradient approximations has to vanish asymptotically.  
\section{Outline of Convergence}
\label{sec:convergence}
We analyze our algorithms using the theory of multiple time-scale stochastic approximation \cite[Chapter 6]{borkar2008stochastic}. Both the algorithms comprise of updates to the policy parameter $\theta$ on the faster time-scale and to the Lagrange multiplier $\lambda$ on the slower time-scale. 
We first provide the analysis for PG-CVaR-SA algorithm and later describe the necessary modification for the mini-batch variant.

Before the main proof, we recall the following well-known result (cf. Chapter 2 of \cite{borkar2008stochastic}) related to convergence of stochastic approximation schemes under the existence of a so-called {\em Lyapunov function}:
\begin{proposition}
\label{prop:lyapunov}
Consider the following recursive scheme:
\begin{align}
\label{eq:sag}
\theta_{n+1} = \theta_n + \gamma_n (F(\theta_n)+\xi_{n+1}),
\end{align}
where $F:\R^d\rightarrow\R^d$ is a $L$-Lipschitz map and $\xi_n$ a square-integrable martingale difference sequence with respect to the filtration $\F_n:= \sigma(\theta_m,\xi_m, m\le n)$. Moreover, $\E[\l \xi_{n+1}\r^2 \mid \F_n] \le K(1+\l \theta_n\r^2)$ for some $K>0$.   The step-sizes $\gamma_n$ satisfy $\sum\limits_{n = 1}^\infty \gamma_n = \infty$ and $\sum\limits_{n = 1}^\infty \gamma_n^2  < \infty$.  

\noindent\textbf{Lyapunov function.} Suppose there exists a continuously differentiable $V: \R^d \rightarrow [0,\infty)$ such that $\lim_{\l \theta \r \rightarrow \infty} V(\theta) = \infty$. Writing $\Z := \{ \theta \in \R^d \mid V(\theta)=0\}\ne \phi$, $V$ satisfies $\left<F(\theta),\nabla V(\theta)\right> \le 0$ with equality if and only if $\theta \in \Z$.   

Then, $\theta_n$ governed by \eqref{eq:sag} converges a.s. to an internally chain transitive set contained in $\Z$.
\end{proposition}
  
The steps involved in proving the convergence of PG-CVaR-SA are as follows:
\subsection*{Step 1: CVaR estimation on fastest time-scale}
Owing to the time-scale separation, $\theta$ and $\lambda$ can be assumed to be constant (quasi-static) while analyzing the VaR/CVaR estimation procedure. 
We first show that the VaR estimate $\xi_n$ converges to the corresponding true value $\var(C^\theta(s^0))$. 
This can inferred by observing that $V$ (see \eqref{eq:vV}) itself serves as the Lyapunov function and the fact that the step-sizes satisfy (A3) implies the iterates remain bounded. Thus, by an application of Proposition \ref{prop:lyapunov}, it is evident that the recursion \eqref{eq:RMVaR} converges to a point in the set $\{ \xi  \mid V(\xi)=0\}$. Since every local minimum is a global minimum for $V$, the iterates $\xi_n$ will converge to $\var(C^\theta(s^0))$. 
Establishing the convergence of the companion recursion in \eqref{eq:RMCVaR} to estimate $\cvar(C^\theta(s^0))$ is easier. This is because \eqref{eq:RMCVaR} is a plain averaging update that uses the VaR estimate $\xi_n$ from \eqref{eq:RMVaR}.

\subsection*{Step 2: Policy update on intermediate time-scale}
We provide the main arguments to show that $\theta_t$ governed by \eqref{eq:pgcvarsa} converges to asymptotically stable equilibrium points of the following ODE:
\begin{align}
\label{eq:theta-ode}
\dot{\theta}_t = \nabla_\theta \L^{\theta_t, \lambda}(s^0) = \nabla_\theta G^{\theta_t}(s^0) + \lambda \nabla_\theta \cvar(C^{\theta_t}(s^0))
\end{align}

Since $\lambda$ is on the slowest timescale, its effect is 'quasi-static' on the $\theta$-recursion. Further, since the CVaR estimation and necessary gradient estimates using likelihood ratios are on the faster timescale, the $\theta$-update in~\eqref{eq:pgcvarsa} views these quantities as almost equilibrated.
Thus, the $\theta$-update in~\eqref{eq:pgcvarsa} can be seen to be asymptotically equivalent to the following in the sense that the difference between the two updates is $o(1)$: 
\begin{align*}
\theta_{t+1} = \theta_t - \gamma_t \left(\nabla_\theta G^{\theta_t}(s^0) + \lambda \nabla_\theta \cvar(C^{\theta_t}(s^0)) \right)\nonumber,
\end{align*}
Thus,~\eqref{eq:pgcvarsa} can be seen to be a discretization of the ODE~\eqref{eq:theta-ode}. Moreover, $\L^{\theta,\lambda}(s^0)$ serves as the Lyapunov function for the above recursion, since $\dfrac{d \L^{\theta,\lambda}(s^0)}{d t}  
= \nabla_\theta \L^{\theta,\lambda}(s^0) \dot \theta
= \nabla_\theta \L^{\theta,\lambda}(s^0) \big(-\nabla_\theta \L^{\theta,\lambda}(s^0)\big) < 0.$
Thus, by an application of Kushner-Clark lemma \cite{kushner-clark}, $\theta$-recursion in \eqref{eq:pgcvarsa} can be seen to converge to the asymptotically stable attractor for the ODE~\eqref{eq:theta-ode} .   

\subsection*{Step 3: Lagrange multiplier update on slowest time-scale}
This is easier in comparison to the other steps and follows using arguments similar to that used for constrained MDPs in general by \cite{borkar2005actor}. The $\lambda$ recursion views $\theta$ as almost equilibrated owing to time-scale separation and converges to the set of asymptotically stable equilibria of the following system of ODEs:
\begin{align}
\label{eq:lambda-ode}
 \dot \lambda_t \;\;=\;\; \check\Gamma_\lambda\big(\nabla_\lambda \L^{\theta^{\lambda_t},\lambda_t}(s^0)\big) \;\;=\;\; \check\Gamma_\lambda\big(\cvar(C^{\theta^{\lambda_t}}(s^0)) - K_\alpha\big)
\end{align}
where $\theta^\lambda$ is the value of the converged policy parameter $\theta$ when multiplier $\lambda$ is used. $\check{\Gamma}_\lambda$ is a suitably defined projection operator that keeps $\lambda_t$ evolving according to \eqref{eq:lambda-ode} bounded. 
Next, the PG-CVaR-SA algorithm converges to the a (local) saddle point of $\L^{\theta,\lambda}(s^0)$, i.e., to a tuple $(\theta^*,\lambda^*)$ that are a local minimum w.r.t.~$\theta$ and a local maximum w.r.t.~$\lambda$ of $\L^{\theta,\lambda}(s^0)$.

The two claims above related to the convergence of $\lambda$-recursion and overall convergence follow using arguments similar to that by \cite{borkar2005actor,Prashanth13AC}. In particular, the former claim follows using standard stochastic approximation by viewing $\lambda$-recursion as performing gradient ascent, whereas the latter claim requires invocation of the envelope theorem of mathematical economics~\cite{mas1995microeconomic}.

\noindent\textbf{PG-CVaR-mB.} The proof for mini-batch variant differs only in the first step, i.e., estimation of VaR/CVaR and necessary gradients. Assume that the number $m_n$ of mini-batch samples used for averaging in \eqref{eq:cvarmb}--\eqref{eq:cvar-grad-mb}), increases with $n$. Thus, a straightforward application of law of large numbers establishes that the empirical mean estimates in \eqref{eq:cvarmb}--\eqref{eq:cvar-grad-mb} converge to their corresponding true values. The rest of the proof follows in a similar manner as PG-CVaR-SA.
\section{Extension to incorporate Importance Sampling}
\label{sec:is}
In this section, we incorporate an importance sampling procedure in the spirit of  \cite{lemaire2010unconstrained,bardou2009computing} to speed up the estimation procedure for VaR/CVaR in our algorithms. \\
\subsection{Importance sampling.}
Given a random variable $X$ with density $p(\cdot)$ and a function $H(\cdot)$, the aim of an IS based scheme is to estimate the expectation $\E(H(X))$ in a manner that reduces the variance of the estimates. Suppose $X$ is sampled using another distribution with density $\tilde p(X,\eta)$ (parameterized by $\eta$), such that $\tilde p(X,\eta)=0 \Rightarrow p(X)=0$ (an absolute continuity condition).
Then, 
\begin{align}
\E(H(X)) = \E\left[H(X)\dfrac{p(X)}{\tilde p(X,\eta)}\right].
\label{eq:isest}
\end{align}
The problem is to choose the parameter $\eta$ of the sampling distribution so as to minimizes the variance of the above estimate.  

A slightly different approach based on mean-translation is taken in a recent method proposed by \cite{lemaire2010unconstrained}.  
 By translation invariance, we have
\begin{align}
\E[H(X)]=\E\left[H(X+\eta)\frac{p(X+\eta)}{p(X)}\right],
\label{eq:translation}
\end{align}
and the objective is to find a $\eta$ that minimizes the following variance term:
\begin{align}
Q(\eta):=\E\left[H^{2}(X+\eta)\frac{p^{2}(X+\eta)}{p^{2}(X)}\right].
\label{eq:Qeta}
\end{align}
If $\nabla Q$ can be written as an expectation, i.e., $\nabla Q(\eta) = \E[ q(\eta,X)]$, then one can hope to estimate this expectation (and hence minimize $Q$) using a stochastic approximation recursion. However, this is not straightforward since 
$\l q(\eta,x) \r$ is required to be sub-linear to ensure convergence of the resulting scheme\footnote{As illustrated by \cite[Section 2.3]{bardou2009computing}, even for a standard Gaussian distributed $X$, i.e., $X\sim\N(0,1)$, the function $q(\eta,x)=\exp(|\eta|^2/2-\eta x)H^2(x) (\eta-x)$  and hence not sub-linear.}.

One can get around this problem by double translation of $\eta$ as suggested first by \cite{lemaire2010unconstrained} and later used by \cite{bardou2009computing} for VaR/CVaR estimation. Formally, under classic log-concavity assumptions on $p(X)$, it can be shown that $Q$ is finite, convex and differentiable, so that
\begin{align}
\nabla Q(\eta):= & \E\left[H(X-\eta)^{2} \frac{p^{2}(X-\eta)}{p(X)p(X-2\eta)}\frac{\nabla p(X-2\eta)}{p(X-2\eta)}\right].
\label{eq:nablaQ}
\end{align}
Writing $K(\eta,X):=\frac{p^{2}(X-\eta)}{p(X)p(X-2\eta)}\frac{\nabla p(X-2\eta)}{p(X-2\eta)}$, one can bound $K(\eta,X)$ by a deterministic function of $\eta$ as follows: $|K(\eta,X)| \leq e^{2\rho|\eta|^{b}}(A|x|^{b-1}+A|\eta|^{b-1}+B)$, for some constants $\rho,A$ and $B$. 
The last piece before present an IS scheme is related to controlling the growth of $H(X)$. We assume that $H(X)$ is controlled by another function $W(X)$ that satisfies
$\forall x, |H(x)| \leq  W(x), W(x+y) \leq C(1+W(x))^{c} (1+W(y))^{c}$ and $\E\left[|X|^{2(b-1)}W(X)^{4c}\right] < \infty.$

An IS scheme based on stochastic approximation updates as follows:
\begin{equation}
\eta_{n}=\eta_{n-1}-\gamma_{n}\tilde q(\eta_{n-1},X_{n}), 
\label{eq:URIS}
\end{equation}
where $\tilde{q}(\eta,X):= H(X-\eta)^{2} e^{-2\rho|\theta|^{b}} K(\eta,X)$.
In lieu of the above discussion, \\$\l\tilde{q}(\eta,X)\r$ can be bounded by a linear function of $\l\eta\r$ and hence, the recursion \eqref{eq:URIS} converges to the set $\left\{\eta \mid \nabla Q(\eta)=0\right\}$ (See Section 2.3 by \cite{bardou2009computing} for more details).
\\[2ex]
\subsection{IS for VaR/CVaR estimation.}
Let $D:=(s_0,a_0,\ldots,s_{\tau-1},a_{\tau-1})$ be the random variable corresponding to an SSP episode and let $D_n:=(s_{n,0},a_{n,0},\ldots,s_{n,\tau-1},a_{n,\tau-1})$ be the $n$th sample simulated using the distribution of $D$.  Recall that the objective is to estimate the VaR/CVaR of the total cost $C^\theta(s^0)$, for a given policy parameter $\theta$ using samples from $D$.   

Applying the IS procedure described above to our setting is not straightforward, as one requires the knowledge of the density, say $p(\cdot)$, of the random variable $D$. Notice that the density $p(D)$ can be written as
$p(D) = \prod\limits_{m=0}^{\tau-1} \pi_\theta(a_m\mid s_m) P(s_{m+1}\mid s_m,a_m).$ As pointed out in earlier works (cf. \cite{sutton1998reinforcement}), the ratio $\frac{p(d)}{p(d')}$ can be computed for two (independent) episodes $d$ and $d'$ without requiring knowledge of the transition dynamics. 

In the following, we use $\tilde p(D_n) :=\prod\limits_{m=0}^{\tau-1} \pi_\theta(a_{n,m}\mid s_{n,m})$ as a proxy for the density $p(D_n)$ and apply the IS scheme described above to reduce the variance of the VaR/CVaR estimation scheme \eqref{eq:RMVaR}--\eqref{eq:RMCVaR}.  The update rule of the resulting recursion is as follows:
\begin{align}
\xi_n & = \xi_{n-1} - \zeta_{n,1} e^{-\rho|\eta|^{b}}\left(1-\frac{1}{1-\alpha} \mbox{ \bf 1}_{\left\{C_n+\eta_{n-1}\geq
\xi_{n-1}\right\}}\frac{\tilde p(D_n+\eta_{n-1})}{\tilde p(D_n)}\right) ,\label{eq:varis} \\
\eta_n & = \eta_{n-1} - \zeta_{n,1} e^{-2 \rho |\eta_{n-1}|^{b}}\mbox{\bf 1}_{\left\{C_n-\eta_{n-1} \geq \xi_{n-1}\right\}} \frac{\tilde p^{2}(D_n-\eta_{n-1})}{\tilde p(D_n)\tilde p(D_n-2\eta)}\frac{\nabla \tilde p(D_n-2\eta_{n-1})}{\tilde p(D_n-2\eta_{n-1})}.\label{eq:varisreduce}\\
\psi_n & = \psi_{n-1} - \zeta_{n,2} \bigg(\psi_{n-1}-\xi_{n-1}-\frac{1}{1-\alpha}(C_n+\mu_{n-1}-\xi_{n-1}) \label{eq:cvaris}\\
& \hspace{12em} \mbox{ \bf 1}_{\left\{C_n+\mu_{n-1} \geq \xi_{n-1} \right\}}\frac{\tilde p(D_n+\mu_{n-1})}{\tilde p(D_n)}\bigg),\nonumber \\
\mu_n & = \mu_{n-1} - \zeta_{n,2} \frac{e^{-2\rho | \mu_{n-1} |^{b}}}{1+W(-\mu_{n-1})^{2c}+\xi_{n-1}^{2}} \left(C_n-\mu_{n-1}-\xi_{n-1}\right)^{2} .\label{eq:cvarisreduce}\\
& \hspace{5em} \times \mbox{\bf 1}_{\left\{C_n-\mu_{n-1} \geq \xi_{n-1} \right\}}\frac{\tilde p^{2}(D_n-\mu_{n-1})}{\tilde p(D_n)\tilde p(D_n-2\mu_{n-1})}\frac{\nabla \tilde p(D_n-2\mu_{n-1})}{\tilde p(D_n-2\mu_{n-1})}.\nonumber
\end{align}  
In the above, \eqref{eq:varis} estimates the VaR, while \eqref{eq:varisreduce} attempts to find the best variance reducer parameter for VaR estimation procedure. 
Similarly, \eqref{eq:cvaris} estimates the CVaR, while \eqref{eq:varisreduce} attempts to find the best variance reducer parameter for CVaR estimation procedure.

\noindent\textbf{Note on convergence.} 
Since we approximated the true density $p(D)$ above using the policy, the convergence analysis of the above scheme is not straightforward.
The difficult part is to establish that one can use the approximation $\tilde p(\cdot)$ in place of the true density $p(\cdot)$. Once this holds, then it can be shown that the tuple $(\eta_n,\mu_n)$ updated according to \eqref{eq:varisreduce} and \eqref{eq:cvarisreduce}, converge to the optimal variance reducers $(\eta^*,\mu^*)$, using arguments similar to that in Proposition 3.1 of \cite{bardou2009computing}. $(\eta^*,\mu^*)$ minimize the convex functions \\$Q_{1}(\eta,\xi^{*}_{\alpha}):= \E\left[\mbox{ \bf 1}_{\left\{C^\theta(s^0) \geq \xi^{*}_{\alpha}\right\}} \frac{p(D)}{p(D-\eta)}\right]$ and \\$Q_{2}(\mu,\xi^{*}_{\alpha}):=\E\left[\left(C^\theta(s^0)-\xi^{*}_{\alpha}\right)^{2}\mbox{\bf1}_{\left\{C^\theta(s^0) \geq \xi^{*}_{\alpha}\right\}} \frac{p(D)}{p(D-\mu)}\right],$ where $\xi^*_\alpha$ is a $\var(C^\theta(s^0))$.



\section{Comparison to previous work}
\label{sec:related-work}
In comparison to~\cite{borkar2010risk} and \cite{tamar2014policy}, which are the most closely related contributions, we would like to point out the following:\\
  \begin{inparaenum}[\bfseries(i)]
   \item The authors by \cite{borkar2010risk} develop an algorithm for a (finite horizon) CVaR constrained MDP, under a separability condition for the single-stage cost. On the other hand, without a separability condition,  we devise policy gradient algorithms in a SSP setting and our algorithms are shown to converge as well; and \\
   \item The authors by \cite{tamar2014policy} derive a likelihood estimate for the gradient of the CVaR of a random variable. However, they do not consider a risk-constrained SSP and instead optimize only CVaR.  In contrast, we employ a convergent procedure for estimating CVaR that is motivated by a well-known convex optimization problem \cite{rockafellar2000optimization} and then employ policy gradients for both the objective and constraints to find a locally risk-optimal policy. 
  \end{inparaenum}

\section{Conclusions}
\label{sec:conclusions}
In this paper, we proposed two novel algorithms to compute a risk-optimal policy in a stochastic shortest path problem. We used Conditional Value-at-Risk (CVaR) as a risk measure and this is motivated by applications in finance and energy markets. 
Both the algorithms incorporated a CVaR estimation procedure along the lines of \cite{bardou2009computing}, which in turn is based on the well-known convex optimization representation by \cite{rockafellar2000optimization}. For the purpose of finding a locally risk-optimal policy, our algorithms employed four tools: stochastic approximation, mini batches, policy gradients and importance sampling. In particular, stochastic approximation or mini-batches are used to approximate CVaR estimates/necessary gradients in the algorithms, while the gradients themselves are obtained using the likelihood ratio technique. Further, since CVaR is an expectation that conditions on the tail probability, to speed up CVaR estimation we incorporated an importance sampling procedure along the lines of \cite{bardou2009computing}. We established asymptotic convergence of both the algorithms. 

There are several future directions to be explored such as 
\begin{inparaenum}[\bfseries(i)] \item obtaining finite-time bounds for our proposed algorithms , \item handling very large state spaces using function approximation, and \item applying our algorithms in practical contexts such as portfolio management in finance/energy sectors and revenue maximization in the re-insurance business. \end{inparaenum}


\bibliographystyle{plainnat}
\bibliography{risk-cvar}

\end{document}